\documentclass{article}

\usepackage[nonatbib, preprint]{neurips_2020_ml4ad}

\usepackage[utf8]{inputenc} % allow utf-8 input
\usepackage[T1]{fontenc}    % use 8-bit T1 fonts
\usepackage{hyperref}       % hyperlinks
\usepackage{url}            % simple URL typesetting
\usepackage{booktabs}       % professional-quality tables
\usepackage{amsfonts}       % blackboard math symbols
\usepackage{nicefrac}       % compact symbols for 1/2, etc.
\usepackage{microtype}      % microtypography
\usepackage{graphicx}
\usepackage{caption}
\usepackage{subcaption}
\usepackage[dvipsnames]{xcolor}
\usepackage{siunitx}
\usepackage{array}
\newcolumntype{C}[1]{>{\centering\arraybackslash}p{#1}}

\title{Multi-Task Network Pruning and Embedded Optimization for Real-time Deployment in ADAS}

\begin{document}

\author{%
  Flora Dellinger\\
  Valeo, France\\
  \texttt{flora.dellinger@valeo.com} \\
  % examples of more authors
  \And
  Thomas Boulay \\
  Valeo, France\\
  \texttt{thomas.boulay@valeo.com} \\
   \And
  Diego Mendoza Barrenechea \\
  Valeo, France\\
  \texttt{diego.mendoza@valeo.com} \\
   \And
  Said El-Hachimi \\
  Valeo, France\\
  \texttt{said.el-hachimi@valeo.com} \\
   \And
  Isabelle Leang\\
  Valeo, France\\
  \texttt{isabelle.leang@valeo.com} \\
   \And
  Fabian B\"urger\\
  Valeo, France\\
  \texttt{fabian.burger@valeo.com} \\
}

\maketitle

\begin{abstract}
Camera-based Deep Learning algorithms are increasingly needed for perception in Automated Driving systems. However, constraints from the automotive industry challenge the deployment of CNNs by imposing embedded systems with limited computational resources. In this paper, we propose an approach to embed a multi-task CNN network under such conditions on a commercial prototype platform, i.e. a low power System on Chip (SoC) processing four surround-view fisheye cameras at $10$ FPS.

The first focus is on designing an efficient and compact multi-task network architecture. Secondly, a pruning method is applied to compress the CNN, helping to reduce the runtime and memory usage by a factor of 2 without lowering the performances significantly. Finally, several embedded optimization techniques such as mixed-quantization format usage and efficient data transfers between different memory areas are proposed to ensure real-time execution and avoid bandwidth bottlenecks.
The approach is evaluated on the hardware platform, considering embedded detection performances, runtime and memory bandwidth.
Unlike most works from the literature that focus on classification task, we aim here to study the effect of pruning and quantization on a compact multi-task network with object detection, semantic segmentation and soiling detection tasks.
\end{abstract}

\section{Introduction} \label{intro}

For \textbf{Advanced Driver-Assistance Systems (ADAS)}, an accurate perception of the surrounding environment is of utmost importance. Accurate localization and proper classification of objects such as cars and pedestrians, as well as road, lane markings and sidewalks are essential for path planning and obstacle avoidance. Image-based perception is used for increasingly complex tasks due to the competitive cost of cameras and the advances in computer vision based on neural networks.

Neural networks, particularly \textbf{Convolutional Neural Networks (CNN)} offer state-of-the-art performance in a wide range of computer vision tasks, such as image classification, object detection and semantic segmentation. Therefore, they are particularly interesting for perception in ADAS. Yet, top performing CNNs on common benchmarks such as ImageNet \cite{deng2009imagenet} and COCO \cite{lin2014microsoft} are usually computationally intensive and require a large amount of memory and power, which makes deployment on embedded systems a challenge, particularly in ADAS. Indeed, for reducing costs and power consumption, automotive manufacturers prefer cheaper low power chips with limited computational resources rather than standard GPUs. Also, ADAS are generally based on a multi-camera system whose information must be processed at a high frame rate with high performance, making the problem even more complex.
Although computational cost can be decreased by reducing the CNN network size and depth, it can negatively impact the performance, which makes embedding CNNs on such devices so difficult.

Based on these specific conditions, we have designed a \textbf{compact multi-task network} that can be embedded on a low power System on Chip (SoC). Three tasks are integrated: object detection, semantic segmentation and soiling detection (see Figure \ref{fig:multi-task}). The soiling detection task is useful in adverse weather conditions by detecting dirt and water on the camera lens.  
Such network is more efficient for embedded applications compared to multiple single task networks, since the use of a single shared encoder reduces significantly the number of operations. 
To fit the hardware computational budget, a large effort was made on network compression and embedded optimization. A \textbf{pruning algorithm} is applied to the network to reduce the FLOPs (floating point operations) by almost a factor of 2 with limited impact on performances. Several \textbf{embedded optimizations} are performed to ensure an efficient execution of the network inside the SoC environment. This is one of the first articles introducing the effect of pruning and quantization on a compact multi-task CNN.

The rest of the paper is structured as follows. Section \ref{SystPres} introduces the overall perception system. Our proposals for a multi-task network and CNN compression are presented respectively in Section \ref{MTL} and \ref{NetComp}. Section \ref{EmbOpt} details our embedded optimization. Section \ref{XPRes} discusses the experimental setup and results. Finally, section \ref{Conclusion} summarizes the paper and provides potential future directions.

\section{Perception system overview}\label{SystPres}

Our perception system performs several operations starting with sensor data rendering, CNN execution and finally CNN output interpretation before translating them into actions for our global automotive system. To reach $10$ frames per second (FPS) on the entire system, a System on Chip (SoC) capable of optimizing the whole video pipeline is required. The SoCs available on the automotive market include, on the same chip, several types of computing units dedicated to the image processing and computer vision operations, in addition to the more commonly found CPU or GPU cores.
Moreover, the manufacturer must compromise and set the number of units on the SoC to the minimum needed to keep the power consumption under $10$W.

\begin{figure}[h]
\centering
\includegraphics[width=\textwidth]{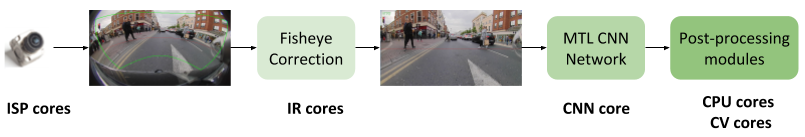}
\caption{Video pipeline on the SoC}
\label{fig:videopipeline}
\end{figure}

Figure \ref{fig:videopipeline} summarizes the video pipeline stages and corresponding units on the SoC.
One or multiple Image Signal Processors (ISPs) transform raw pixels captured by four surround-view $2$ MPixel fisheye cameras into high quality images. ISPs are part of the global image processing block used for demosaicing, noise reduction or lens imperfection correction on the raw signal. A cylindrical fisheye correction is performed by a dedicated Image Rendering (IR) core to reduce distortion before feeding the image to the CNN.

The MTL CNN is executed on a dedicated CNN core which performs convolutions of kernel size $5\times5$ on $4$ inputs and $8$ output channels in a single clock-cycle, reaching about $1$ Tera operations per second (TOPS). Finally, the outputs of the MTL network are post-processed by CPU cores or other computer vision cores (CV cores) available on the SoC. From a global automotive system perspective, the post-processed outputs of the CNN are then fused with several other outputs from other sensors and are finally translated into active safety actions.\\

Consequently, this perception system configuration imposes some constraints to the architecture of the multi-task CNN. Based on these specifications, the number of FLOPs must be limited to $3.5$ GFlops to ensure a deployment on the embedded hardware at $10$ FPS. Starting from these requirements, we will present our approach to design and embed such a multi-task network in the following sections.

\section{Multi-task CNN design}\label{MTL}

\subsection{Related work}\label{sub:MTLStofArt}

\textbf{Object Detection}
State-of-the-art CNNs for object detection are based on a two-stage approach, with region proposals followed by a detection module \cite{ren2015faster}. However, faster algorithms have been proposed based on a single step approach \cite{redmon2016you} \cite{redmon2017yolo9000} \cite{liu2016ssd}, which are more adapted to real-time applications.

\textbf{Semantic Segmentation} State-of-the-art approaches to semantic segmentation trained end-to-end are based on Fully Convolutional Networks \cite{long2015fully} using transposed convolutions \cite{zeiler2010deconvolutional} to recover the spatial information needed for pixel-level classification \cite{ronneberger2015u}.

\textbf{Soiling detection} Soiling detection \cite{uvrivcavr2019soilingnet} can be approached as a coarse semantic segmentation task, based on a $4\times4$ grid for localization as described in \cite{das2020tiledsoilingnet}. 

\textbf{Multi-task learning} For problems where multiple tasks are required, such as object detection and semantic segmentation in autonomous driving applications, using a single neural network with a common encoder and multiple heads presents many benefits. Low-level features can be useful for multiple tasks, and multi-task training can provide a regularization that improves the test set performance \cite{teichmann2018multinet}. Moreover, a common encoder requires less memory and computational resources.

\subsection{Our approach: Overview of 3-task CNN}\label{3tasks}

\begin{figure*}[!t]
\centering
\includegraphics[width=\textwidth]{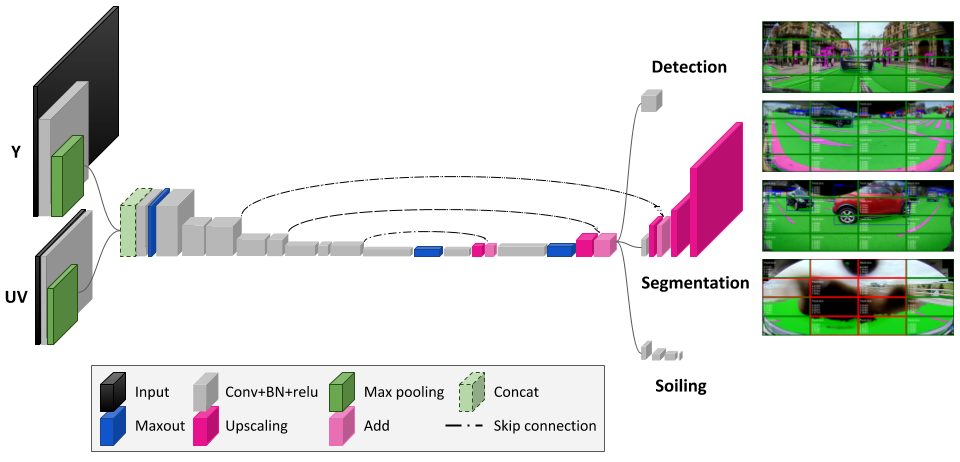}
\caption{Multi-task visual perception network architecture with Y (gray) and UV (color) input}
\label{fig:multi-task}
\end{figure*}

Our 3-task CNN as shown in Figure \ref{fig:multi-task} is composed of a U-Net-like encoder optimized by a NAS (Neural Architecture Search) algorithm based on Evolution Strategies \cite{burger2016understanding}. Particularly the network depth, the number of filters and the position of skip connections and other layers have been optimized to boosted performance while limiting the network complexity. We used a YOLO-like decoder for object detection, a FCN-like decoder for semantic segmentation and for soiling we used the decoder described in \cite{das2020tiledsoilingnet}. The kernel size was set to $5\times5$, which has an efficient implementation on the used embedded system.
Given the computational cost of each convolutional layer, it is important to have as few layers as possible. As a consequence, initial experiments with a limited number of layers could hardly detect objects in the near-field which often cover large parts of the image. To overcome this, we determined a minimal number of convolutional layers required so that each task head output has a receptive field larger than the input image size, based on receptive field computation methods described in \cite{araujo2019computing}.

While compact, this network still requires too much computation for our embedded setup ($6$ GFLOPs while the maximal limit of our setup is $3.5$). The next section will introduce the concepts of pruning and quantization for network compression, followed by our chosen pruning approach for runtime reduction.

\section{Network compression} \label{NetComp}

\subsection{Related work}\label{NetCompStofArt}

Network compression techniques \cite{dally2015, han2016}, such as quantization and pruning, have gained a lot of interest in recent years in order to be able to use CNNs on edge devices for real time applications.

\textbf{Quantization} Low-power hardware use fixed-point precision operations to execute the network. Therefore, a step of quantization \cite{krishna2018} is mandatory to deploy the  network trained with floating-point precision. Quantization is also a way to reduce network size, by representing weights and activations of the model with lower-precision numerical formats (commonly 8/16-bit integers instead of 32-bit floating point). The work of \cite{lai2017deep,lin2016fixed,7178146} proposed post-training quantization methods to efficiently represent the information of the floating-point numbers during training into fixed-point integers. This technique is widely used to embed  single-task CNNs for storage, bandwidth and runtime reduction. However, the impact of quantization on multi-task CNNs has rarely been studied.

\textbf{Pruning} is a compression technique that removes less important connections in a neural network. The idea behind is that neural networks are over-parametrized and thus some neurons don't contribute much to the final result. Starting from a large network with a good performance, the goal is to remove neurons with few contribution in order to obtain a smaller network with similar accuracy \cite{blalock2020state}. Like quantization, this is useful to reduce network storage, bandwidth and inference runtime.

\begin{figure}[t]
\begin{minipage}{.53\textwidth}
    \centering
    \includegraphics[width=\textwidth]{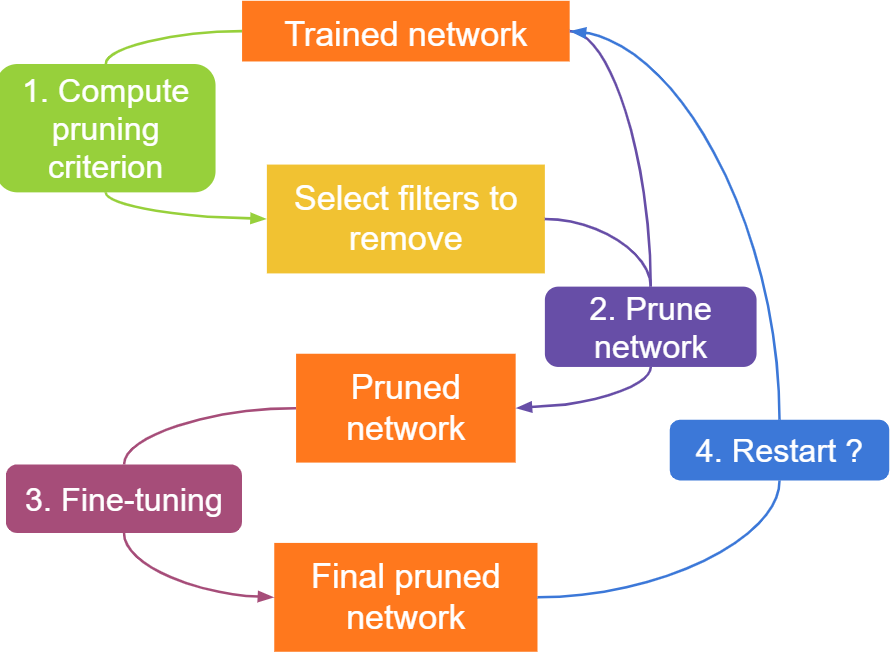}
    \captionof{figure}{Steps of pruning methods.}
    \label{fig:pruning_steps}
\end{minipage}%
\begin{minipage}{.47\textwidth}
    \centering
    \includegraphics[width=0.9\textwidth]{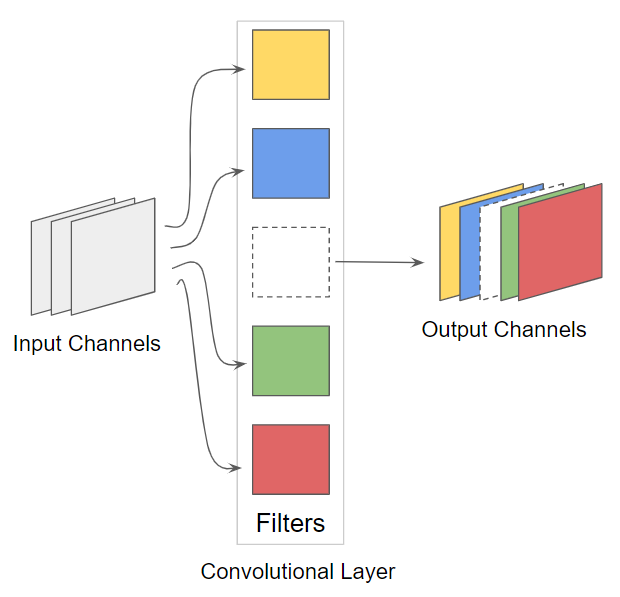}
    \captionof{figure}{Example of filter pruning.}
    \label{fig:filter_pruning}
\end{minipage}
\end{figure}

Pruning has proven its effectiveness through a large number of methods in literature. While it is difficult to compare them due to a lack of benchmark datasets \cite{blalock2020state}, nearly all follow the same steps introduced in \cite{han2016} and stated in Figure \ref{fig:pruning_steps}. That being said, pruning methods can be differentiated based on their pruning granularity, their pruning criterion and their scheduling.

The \textbf{pruning granularity} defines the type of structure pruned in the network.
\begin{itemize}
    \item \textbf{Weight pruning} \textit{(unstructured pruning)} \cite{han2016} consists in inducing sparsity in weights and activations of a model. Weights are transformed into sparse matrices, by setting some elements to zeros. This is effective to reduce the number of parameters and thus network storage. However, most hardware and libraries do not handle sparse convolutions, thus this won't entail a runtime reduction. For a vast majority of hardware and applications, this solution is not helpful.
    \item \textbf{Filter pruning} \textit{(structured pruning)} \cite{li2016}, on the other hand, allows to prune entire filters or channels (see Figure \ref{fig:filter_pruning}). A direct reduction of the FLOPs is thus obtained, leading to a decrease of inference runtime. However, accuracy can strongly be reduced and fine-tuning is usually needed to recover the performance drop.
\end{itemize}

The \textbf{pruning criterion} allows to compare and choose which elements to prune. A score is assigned to each weight or filter to assess its contribution to the network. Scores are then compared locally (within a layer) or globally (across all layers), and elements with scores below a threshold are pruned. Choosing the right criterion is key in pruning success. Various methods have been investigated:
\begin{itemize}
    \item \textbf{Data-dependent criteria} are related to some statistics on a dataset, like the average or percentage of zeros among activations \cite{hu2016network, molchanov2016}.
    \item \textbf{Data-independent criteria} are based on the state of the network: $l1$-norm on filters \cite{li2016}, Taylor expansion \cite{molchanov2016} or scaling factors from Batch Normalization \cite{liu2017slimming}.
\end{itemize}

The \textbf{pruning scheduling:} Some methods conduct one-shot pruning  by compressing the network in a single step \cite{liu2017slimming}, while others prefer to remove a percentage of the network in an iterative manner \cite{han2016}. Most methods carry out some fine-tuning after each pruning step to recover performances.

As stated in \cite{blalock2020state}, there is no perfect pruning method. Results will be different regarding the network size and amount of pruning. Choosing the right method will depend on the type of network, the compression goal and the hardware implementation. We can also note that most methods have only been applied to image classification, which is not representative of the tasks used for CNN. There are only few publications on the use of pruning on other tasks, like semantic segmentation in \cite{chen2020multi} or visual attribute prediction in \cite{he2019}. It is thus necessary to study the impact of pruning on compact networks with complex and multiple visual tasks.

\subsection{Our approach: Pruning of 3-task CNN}\label{mtlpruning}

We have chosen a filter pruning method as our target hardware is optimized for dense convolutions. Two pruning criteria common in the literature are compared: FilterNorm based on the $l1$-norm of weights \cite{li2016} and BatchNorm based on the $\gamma$ scaling factor \cite{liu2017slimming}. Since we have observed that some layers (especially deeper layers) are more pruned than others, filter scores are then compared globally.

The steps of the pruning method are the same as in Figure \ref{fig:pruning_steps}: starting from a pre-trained model, the model is iteratively pruned and fine-tuned until the target runtime is achieved. When the network has reached the desired compression, a longer fine-tuning is carried out to aim for best performances. A trade-off is needed between the amount of network pruned at each step and the fine-tuning length. Experiments have shown that longer fine-tuning with a large compression rate should be preferred over a small compression rate with short fine-tuning. An iterative pruning is necessary here since a high amount of the network needs to be trimmed. A one-shot pruning is too dramatic for accuracy in our case, however, it can be applicable for small amounts of pruning.

After compression, the 3-task CNN can be executed with a limited number of FLOPs. However, some embedded optimizations are still required to fit the target in our case.

\section{Embedded optimization} \label{EmbOpt}

An embedded CNN deployment and optimization for a real product always start with a clear definition of the target to achieve. It must be delineated by taking into account the full system requirements (CNN is usually not the only algorithm running on the SoC). Generally, the most important metrics to track module by module on the hardware are the runtime, the DDR bandwidth and the DDR footprint, DDR (Double Data Rate) being commonly the memory resource shared by all algorithms. For our application, the budget allocated to the CNN is defined based on these $3$ metrics. First, the CNN must be executed to reach $10$ FPS (a frame is defined as a batch of $4$ images coming from our surround-view fisheye cameras). Secondly, taking into consideration the full system bandwidth and other algorithms running in parallel on the CNN, the bandwidth allocated to our CNN module is fixed to $1$ Giga Bytes per second (GBps). In concrete terms, it means that the throughput measured on the SoC for the CNN module should always be less than $1$GBps. Finally, to work on the SoC, the CNN module needs an amount of allocated memory on the DDR called DDR memory footprint or DDR peak usage. For our application, the DDR memory footprint for the CNN part ($4$ cameras) is limited to $14$MB.

\begin{figure}[t]
\centering
\includegraphics[width=0.6\textwidth]{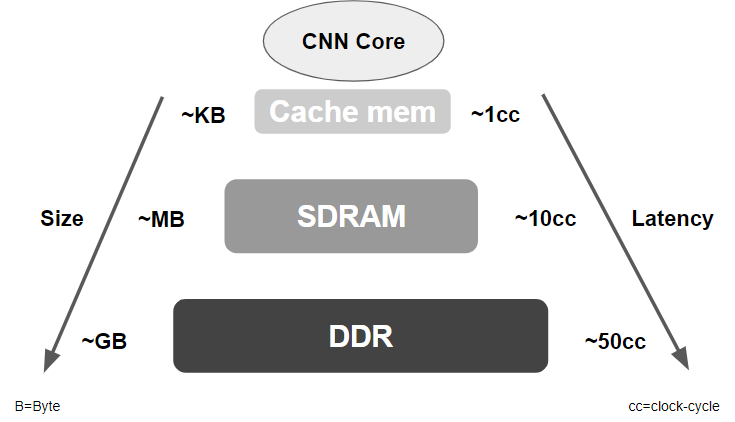}
\caption{Memory areas for the CNN core.}
\label{fig:memareas}
\end{figure}

\subsection{Network optimization}

Network optimization for an embedded deployment often starts as soon as the network architecture is being developed. The paper \cite{boulay2019yuvmultinet} describes several important network architecture optimizations like input/convolution layer shape optimization or layer sequence optimization. The optimizations described in this paper have all been applied to our network enabling to reduce drastically the runtime and the throughput of the network. For example, a YUV input layer has been used instead of the common RGB input layer to reduce by a factor of two the memory transfers needed to process the network's input data.

Once the network architecture is frozen, the filter pruning technique enables a very considerable reduction of the runtime and the bandwidth. A convolutional layer calculation is subdivided into a series of small calculations on the SoC. The spatial dimension is processed by sliding a $5\times5$ window on the feature maps and in the depth dimension only a small group of channels is handled in parallel. One CNN core execution convolves $4$ input channels by $8$ filters to produce $8$ output channels. The layer computation is completed when all the channels (depth dimension) and pixels (spatial dimension) were handled. Hence by definition, when the number of the input/output channels is reduced, the number of CNN core runs and the amount of data to transfer to the CNN core are reduced. Consequently, the runtime and bandwidth are cut down. Furthermore, pruning in early layers of the encoder leads to the strongest reductions since these layers handle high resolution feature maps that require a lot of CNN core executions and data transfers. Table \ref{tab:EmbeddedResults} shows the achieved optimization results.

\subsection{Efficient data transfers}\label{subsub:datatransfer}

A well-known challenge to run optimally an embedded CNN is to efficiently arrange the data for triggering the computation on the core as fast as possible. When the CNN core ends its current calculation, the goal is to reduce as much as possible the waiting time or latency before launching the next calculation. On the SoC, several memory areas are coexisting. The closer the memory area is to the computation core, the lower the latency is, but the lower the size is, too (see Figure \ref{fig:memareas}). The DDR memory area is usually the biggest area on most of the SoCs. Consequently, it is also a memory area shared by multiple algorithms which can run simultaneously. Hence, our optimization for this application is mainly focused on reducing the DDR memory traffic by cleverly arranging the data along the CNN layer executions. The main idea behind this optimization is to avoid as much as possible to transfer the CNN core output data into DDR. This principle can be applied at two levels: inside a layer when the intermediate outputs have to be stored to be accumulated with the next convolution channel outputs and across the layers when the final result of a layer is saved to be reuse as input of the next layer. We worked on these two levels by building chains of CNN core execution nodes. These chains have been built in order to keep the data inside the SDRAM memory. So, we alternate two types of execution scheme according to the remaining memory space in SDRAM. A common scheme of tile execution (horizontal/spatial execution) when SDRAM cannot contain the intermediate results and a vertical scheme where a tile of data is processed through all layers in the chain before moving to the next spatial tile and saving in DDR memory. The effects of these optimizations are presented in Table \ref{tab:EmbeddedResults}.

\subsection{Mixed-quantization optimization}\label{subsub:Mixquant}

The CNN core used for our application is hard-wired for executing $16$-bit operations. Consequently, the opportunity to reduce the bit-width is not offered by the hardware. So, we cannot take advantages of approaches like \cite{wang2019haq, yang2020fracbits} because the operations supported by our hardware accelerator are only 16-bit operations. However, outside the CNN core, a reduction of the bit-width can be considered for reducing the amount of transferred data, especially from and to the DDR memory. Therefore, we propose a $8$-bit/$16$-bit mixed-quantization scheme for our network to reach the embedded target in regards of DDR bandwidth and DDR footprint. The principle is to identify the layers with a high data consumption from and to the DDR and quantify the input and output feature maps of these layers in $8$-bit. Following this scheme, 8-bit instead of $16$-bit numbers are transported around the memory areas until they are used by the CNN core which can cast the numbers into the hard-wired bit-width of $16$-bit. The impact of this optimization is described in Table \ref{tab:EmbeddedResults} regarding the embedded metrics and in Table \ref{tab:KPIs_embedded} regarding the algorithm performances. 

\section{Experiments and results} \label{XPRes}

\subsection{Training and pruning protocol}\label{sub:Protocol}
We run our experiments on the Woodscape autonomous driving dataset \cite{yogamani2019woodscape} using a split of 80/10/10 for train/val/test and a reduced number of classes: road, lane marking, curb, vehicle, two wheeler and person. The joint training policy consists of minimizing a multi-task loss which is an arithmetic weighted sum of individual task losses $L = \sum_{i} w_i L_i$, $i$ representing a task. Task balancing is performed by a static weighting whose values have been optimized by automatic tuning described in \cite{leang2020dynamic}. Each batch includes images with annotations for the 3 tasks. The loss is backpropagated only through the relevant task heads and to the common encoder. The network is trained end-to-end using the Adam optimizer \cite{kingma2014adam}. Our data augmentation policy includes modifications in brightness, contrast, saturation, hue, noise, scaling, spatial translation and horizontal flip.
Our main metrics are mean average precision (mAP) for object detection, mean intersection over union (mIoU) for semantic segmentation and a tile-level mean F1-score for soiling detection. We apply a L2 regularization of $10^{-4}$ and train the network for $64$ epochs.
To assess the impact of using a multi-task network, we have trained three independent single-task networks under the same conditions.

After training, the network is pruned in an iterative manner. At each step, an amount of the network is trimmed to reduce the number of GFLOPs and a fine-tuning of $32$ epochs is carried out on the intermediate network. The pruning stops when the number of GFLOPs is below the embedded target (here $3.5$ GFLOPs). The final model is fine-tuned for $64$ epochs. Two pruning experiments are conducted with the two studied pruning criteria.

\subsection{Multi-task versus single-task networks}

Table \ref{tab:KPIs} summarizes metric values for single-task and multi-task networks after training.
Metrics obtained by single-task networks are all higher than corresponding ones from the multi-task network. However, the gap is limited for object detection ($+0.26$) and semantic segmentation ($+1.78$). Only soiling detection benefits strongly of this single training ($+14.28$), even though performances are still competitive with joint training ($79.78$ for F1-score). Moreover, using single-task networks would multiply the number of FLOPS by three in comparison to the multi-task network ($17.8$ instead of $6.0$ GLOPS).

Choosing a multi-task network helps us then to reach competitive performances with limited computation time.

\begin{table}[t]
\begin{tabular}{|l|C{1.25cm}|C{1.7cm}|C{1.3cm}|c|c|}
\hline
\textbf{\small{Network version}} & \textbf{\small{Detection  \hspace{2cm} mAP}} & \textbf{\small{Segmentation  \hspace{2cm}  mIoU}} & \textbf{\small{Soiling  \hspace{2cm} F1-score}}  & \textbf{\small{Parameters}} & \textbf{\small{GFLOPs}} \\
\hline \hline
\small{Object detection} & \small{$70.75$} & - & - & \small{$2,490,700$} & \small{$5.9$} \\ \hline
\small{Semantic segmentation} & - & \small{$65.20$} & - & \small{$2,511,320$} & \small{$6.0$} \\ \hline
\small{Soiling detection} & - & - & \small{$94.06$} & \small{$2,515,028$} & \small{$5.9$} \\ \hline
\small{Multi-task unpruned} & \small{$70.49$} & \small{$63.42$} & \small{$79.78$} & \small{$2,540,208$} & \small{$6.0$} \\ \hline
\small{Multi-task pruned with FilterNorm} & \small{$68.17$} & \small{$62.64$} & \small{$79.97$} & \small{$741,956$} & \small{$3.5$} \\ \hline
\small{Multi-task pruned with FilterNorm,} & \small{$65.15$} & \small{$61.74$} & \small{$78.90$} & \small{$741,956$} & \small{$3.5$} \\
\small{re-trained from scratch} & & & & & \\ \hline
\end{tabular}
\caption{Performance comparison between single-task and multi-task networks, before and after pruning. The evaluation set contains $5$K (detection/segmentation) and $18$K images (soiling).}
\label{tab:KPIs}
\end{table}

\begin{figure}[t]
    \centering
    \begin{subfigure}[b]{0.49\textwidth}
        \includegraphics[width=\textwidth]{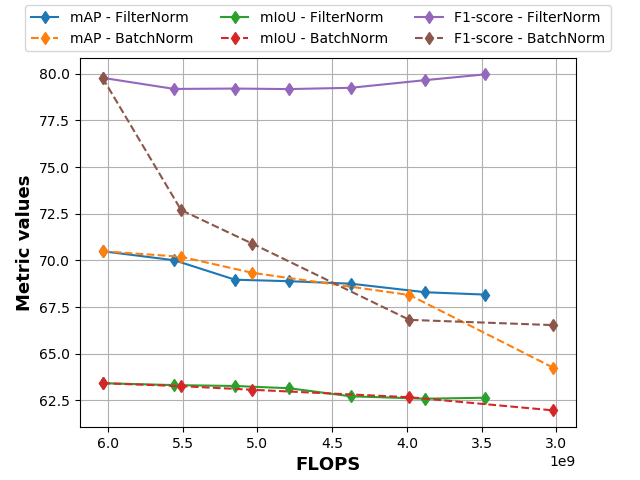}
        \caption{Evolution of metrics over multi-task pruning for two criteria : FilterNorm ($l1$-norm) and BatchNorm ($\gamma$ scaling factor).}
        \label{fig:metrics_pruning}
    \end{subfigure}
    \hfill
    \begin{subfigure}[b]{0.49\textwidth}
    \centering
        \includegraphics[width=\textwidth]{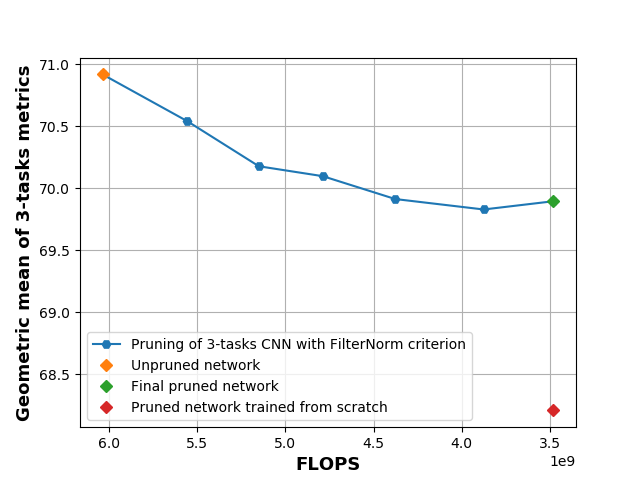}
        \caption{Interest of pruning and comparison with the final pruned network trained from scratch.}
        \label{fig:pruning_interest}
    \end{subfigure}
    \caption{Influence of pruning on multi-task networks. For both figures, the curves display values from the unpruned network (far left) to the final pruned network (far right). In between, there are intermediate networks.}
    \label{fig:metrics}
\end{figure}

\subsection{Network compression experiments}\label{sub:NetCompXP}
    
Figure \ref{fig:metrics} displays the multi-task network behavior over pruning. The evolution of metrics for the 3 tasks are depicted in Figure \ref{fig:metrics_pruning} for the two pruning experiments. While object detection and semantic segmentation metrics behave in a similar way for both criteria (at least until $4.0$ GFLOPS), pruning with BatchNorm criterion leads to a strong degradation of the soiling metric. Thus, the FilterNorm criterion seems more suited to this network and later experiments and discussions will be conducted with this criterion.

Figure \ref{fig:pruning_interest} provides an overview of the network accuracy over FilterNorm pruning via the geometric mean of the 3 metrics, while Table \ref{tab:KPIs} gathers detailed metric values. 
Iterative pruning with FilterNorm allows to strongly reduce the number of FLOPs and parameters, by respectively $42\%$ and $71\%$, with minimum effect on performances. The object detection task ($-2.3\%$ on mAP) is more impacted than semantic segmentation ($-0.78\%$ on mIoU), while the soiling task is even slightly improved ($+0.19\%$ on F1-score).

To assess the utility of pruning compared to training a more compact network directly, the final pruned network has been retrained from scratch on $64$ epochs. For all three tasks, the values are below the ones obtained with pruning ($-3.02\%$, $-0.90\%$ and $-1.07\%$, respectively on the mAP, mIoU, F1-score metrics).
Consequently, pruning is useful and efficient to compress the multi-task CNN. While a small degradation of performances may be observed, the impact is less important on soiling and segmentation tasks than on object detection.

\subsection{Embedded optimization results}\label{sub:EmbOptRes}

Table \ref{tab:EmbeddedResults} presents the embedded optimization impact on the embedded metrics. The impact of pruning is very significant on the $3$ metrics. The pruned network offers a runtime and memory footprint reduction by almost a factor of $2$ compared to the unpruned network. The DDR bandwidth and footprint are respectively reduced by $25\%$ and $4\%$ by the data transfer optimized network (see Section \ref{subsub:datatransfer}). However, a side effect of using the low-latency memory is that smaller tiles (block of pixels loaded in memory) have to be defined to process the network in a smaller amount of memory. It leads to more CNN-core runs and consequently to lower runtime performances. Finally, with the mixed-precision network DDR bandwidth and footprint are reduced and overcome respectively the $1$GBps and $14$MB targets. The runtime also takes advantage of this optimization with a gain of $5\%$ which offers some room on the $10$ FPS goal. In terms of performances (see Table \ref{tab:KPIs_embedded}), the mixed-precision network very slightly penalizes the $3$ task performances compared to the float32 network ($-0.6\%$, $-0.06\%$ and $-3\%$ respectively on the mAP, mIoU, F1-score metrics). This small degradation is a good trade-off in regards of the gain observed on the embedded metrics.

\begin{table}
\begin{tabular}{l|c|c|c|}
\cline{2-4}
        & \textbf{\small{Runtime (FPS)}}     & \textbf{\small{Bandwidth (GBps)}}   & \textbf{\small{Footprint (MB)}}     \\ \cline{2-4} 
        & \multicolumn{3}{c|}{\textbf{\small{Goal}}}                                                     \\ \cline{2-4} 
        & \small{10  }                       & \small{1}                           & \small{14}                      \\ \hline
\multicolumn{1}{|l|}{\textbf{\small{Network version}}}         & \multicolumn{3}{c|}{\small{\textbf{Measured}}}                                                 \\ \hline
\multicolumn{1}{|l|}{\small{Unpruned network}}                   & \small{8} & \small{1.6} & \small{30} \\ \hline
\multicolumn{1}{|l|}{\small{Pruned network}}                   & \small{14} & \small{1.2} & \small{15} \\ \hline
\multicolumn{1}{|l|}{\small{Data transfer optimized network}} & \small{10} & \small{1} & \small{14.4} \\ \hline
\multicolumn{1}{|l|}{\small{Mixed-precision network}}          & \small{10.5} & \small{0.96} & \small{13.2} \\ \hline
\end{tabular}
\caption{Embedded optimization results.}
\label{tab:EmbeddedResults}
\end{table}

\begin{table}
\begin{tabular}{|l|C{1.3cm}|C{2.2cm}|C{1.3cm}|c|c|}
\hline
\textbf{\small{Network version}} & \textbf{\small{Detection  \hspace{2cm} mAP}} & \textbf{\small{Segmentation  \hspace{2cm}  mIoU}} & \textbf{\small{Soiling  \hspace{2cm} F1-score}}  & \textbf{\small{Parameters}} & \textbf{\small{GFLOPs}} \\
\hline \hline
\small{float32 precision} & \small{$68.17$} & \small{$62.64$} & \small{$79.97$} & \small{$741,956$} & \small{$3.5$} \\ \hline
\small{16-bit precision} & \small{$67.85$} & \small{$62.64$} & \small{$78.68$} & \small{$741,956$} & \small{N.A} \\ \hline
\small{8-bit/16-bit mixed-precision} & \small{$67.76$} & \small{$62.60$} & \small{$77.33$} & \small{$741,956$} & \small{N.A} \\ \hline
\end{tabular}
\caption{Performance comparison of different embedded versions of our pruned multi-task network. The evaluation set contains $5$K (detection/segmentation) and $18$K images (soiling).}
\label{tab:KPIs_embedded}
\end{table}

\section{Conclusion} \label{Conclusion}
In this paper, we presented a series of algorithm and embedded optimization techniques necessary for deploying a CNN capable of processing three perception tasks in parallel with reasonable performance on a low power SoC at 10 FPS for four surround-view cameras in a vehicle. First of all, we designed an initial multi-task network with a NAS-based architecture considering a compromise between the hardware constraints and perception performances. Then, we proposed an iterative filter pruning method to greatly compress the network reducing the runtime execution by a factor of 2 without significant performance loss. Finally, we proposed several embedded optimization techniques to achieve real-time execution on the SoC such as efficient data transfers and mixed 8/16-bit quantization for reducing DDR bandwidth and footprint. This is one of the first papers presenting pruning and quantization results on a compact multi-task network for automotive applications.

For future work, we would like to explore other pruning methods to improve the network performance, such as the lottery ticket hypothesis \cite{frankle2018lottery} or the automated gradual pruning \cite{zhu2017prune}. We also would like to study the impact of such compression on network robustness \cite{gui2019} and thus safety for automotive applications. To further optimize the network quantization step, the study of other calibration methods would be useful for minimizing the performance degradation even more.

% \begin{ack}

% The authors would like to thank their employer for the opportunity to work on fundamental research. We would also like to thank for reviewing the paper and providing feedback.

% \end{ack}

\bibliographystyle{plain}
\medskip

\small

% [1] Alexander, J.A.\ \& Mozer, M.C.\ (1995) Template-based algorithms for
% connectionist rule extraction. In G.\ Tesauro, D.S.\ Touretzky and T.K.\ Leen
% (eds.), {\it Advances in Neural Information Processing Systems 7},
% pp.\ 609--616. Cambridge, MA: MIT Press.

% [2] Bower, J.M.\ \& Beeman, D.\ (1995) {\it The Book of GENESIS: Exploring
%   Realistic Neural Models with the GEneral NEural SImulation System.}  New York:
% TELOS/Springer--Verlag.

% [3] Hasselmo, M.E., Schnell, E.\ \& Barkai, E.\ (1995) Dynamics of learning and
% recall at excitatory recurrent synapses and cholinergic modulation in rat
% hippocampal region CA3. {\it Journal of Neuroscience} {\bf 15}(7):5249-5262.

\bibliography{neurips_2020_ml4ad}

\end{document}